\documentclass[11pt]{article}
\usepackage{graphicx,amsmath,amsfonts,amssymb,fullpage}
\usepackage[margin=1.0in]{geometry}
\usepackage[natbibapa]{apacite}
\usepackage{caption}
\usepackage{setspace}
\usepackage{algorithmic}
\usepackage{url}
\newcommand{\hi}[1]{^{(#1)}}

\newcommand{\amax}[1]{\underset{#1}{{\text{argmax\ }}}}

\newcommand{\tr}[0]{\text{tr}}

\newcommand{\existsexactly}[0]{\exists !}

\begin{document}

\title{\textbf{The Emergence of Organizing Structure in Conceptual Representation}}
\date{}
\author{Brenden M. Lake,$^{1,2}$ Neil D. Lawrence,$^3$ Joshua B. Tenenbaum,$^{4,5}$ \\
$^1$Center for Data Science, New York University \\
$^2$Department of Psychology, New York University \\
$^3$Department of Computer Science, University of Sheffield\footnote{This work was completed before N. Lawrence joined Amazon Research Cambridge.} \\
$^4$Department of Brain and Cognitive Sciences, Massachusetts Institute of Technology \\
$^5$Center for Brains Minds and Machines
}

\maketitle
\begin{abstract}
Both scientists and children make important structural discoveries, yet their computational underpinnings are not well understood. Structure discovery has previously been formalized as probabilistic inference about the right structural form --- where form could be a tree, ring, chain, grid, etc. [Kemp \& Tenenbaum (2008). The discovery of structural form. \textit{PNAS}, 105(3), 10687-10692]. Although this approach can learn intuitive organizations, including a tree for animals and a ring for the color circle, it assumes a strong inductive bias that considers only these particular forms, and each form is explicitly provided as initial knowledge. Here we introduce a new computational model of how organizing structure can be discovered, utilizing a broad hypothesis space with a preference for sparse connectivity. Given that the inductive bias is more general, the model's initial knowledge shows little qualitative resemblance to some of the discoveries it supports. As a consequence, the model can also learn complex structures for domains that lack intuitive description, as well as predict human property induction judgments without explicit structural forms. By allowing form to emerge from sparsity, our approach clarifies how both the richness and flexibility of human conceptual organization can coexist.
\end{abstract}

\section{Introduction}

Structural discoveries play an important role in science and cognitive development \citep{Carey2009,Kuhn1962}. In biology, Linnaeus realized that living things were best organized as a tree, displacing the ``great chain of being'' used for centuries before. Modern chemistry began with Mendeleev's discovery of the periodic structure of elements. In cognitive development, children realize that the days of the week form a cycle and that social networks form cliques. Children do not initially treat comparative relations such as ``longer than'' as transitive, but they realize the unidimensional structure around age 7 or 8 \citep{Inhelder1964}. While learning the names for objects, children tend to learn just one name per object, but they must later discover that names can be organized in a taxonomy \citep{Markman1989}. Yet, other forms of learning are not marked by clear structural insights and transitions. Everyday forms of learning, such as a child acquiring facts about animals and artifacts, have been characterized as accumulating correlations and gradually differentiating concepts \citep{RogersMcClelland2004}. What distinguishes more ordinary forms of learning from genuine structural discoveries?

This distinction is not well understood computationally. In machine learning, many standard algorithms learn only a single form of structure that is decided on in advance -- whether it is learning clusters (e.g., k-means), trees (hierarchical clustering), or multidimensional spaces (principal component analysis). It is usually up to the human practitioner, rather than the algorithms themselves, to decide which type of structure is most appropriate for the data. Other approaches such as neural networks \citep[deep learning; ][]{Lecun2015} make very general representational assumptions and can model data with many different structural characteristics. However, the learned structure is only implicit in the connection weights, with no obvious mechanism for differentiating between genuine structural discoveries from other forms of learning.

\begin{figure}
\centering
\includegraphics[width=6.5in]{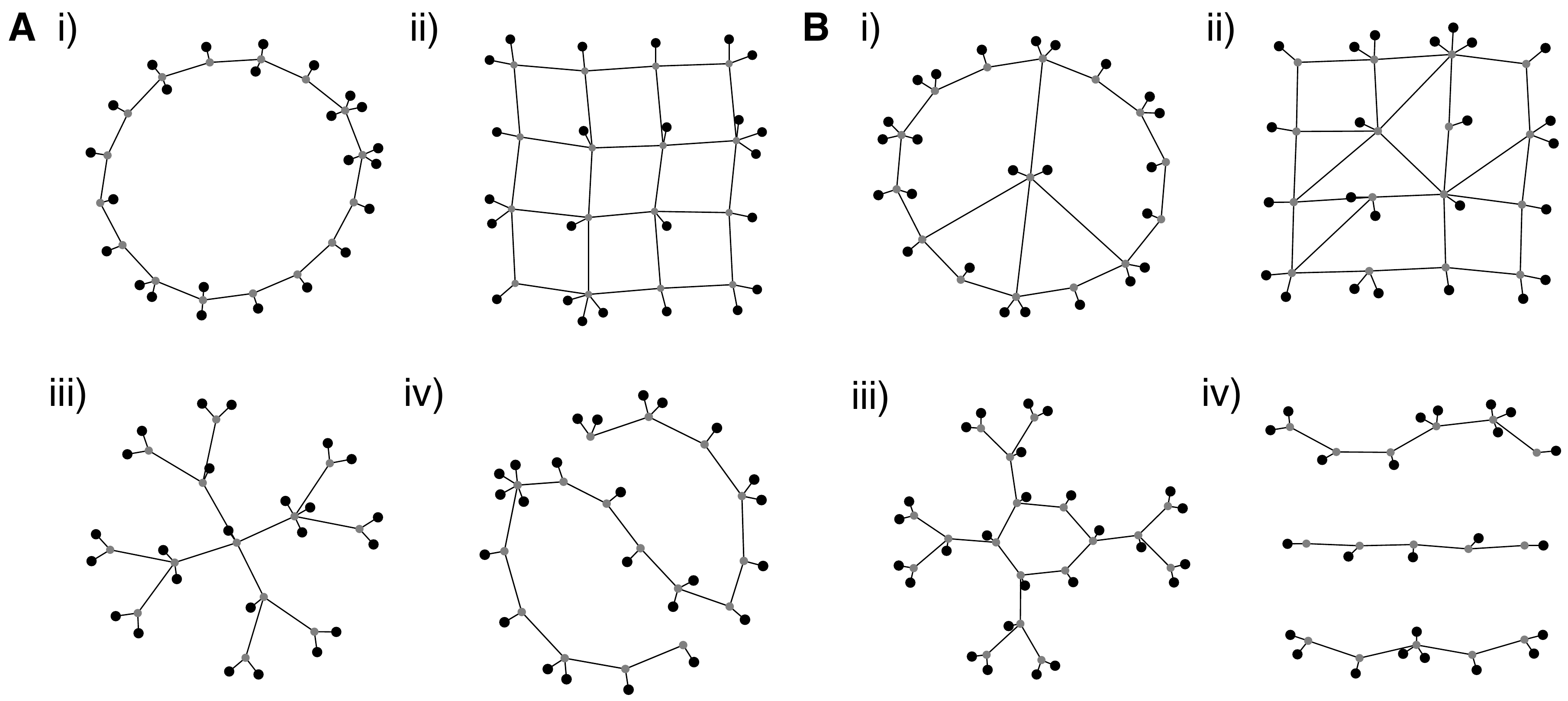}
\caption{Structures that can be represented (A; i: ring, ii: grid, iii: tree, iv: chain), or cannot be represented (B), by the structural forms of Kemp and Tenenbaum \citep{Kemp2008}. All of the structures (A \& B) are sparse, meaning they have relatively few edges (lines) between the nodes. Sparsity is the foundation of the computational model introduced in this work.
\label{fig_synthdata}}
\end{figure}

Previous work from our group investigated how the form of structure can be learned from data and represented explicitly \citep{Kemp2008}. Structure selection can be formalized as probabilistic inference over a set of discrete and mutually exclusive hypotheses such as rings (i), grids (ii), trees (iii), chains (iv), etc. called \emph{structural forms} (Fig. \ref{fig_synthdata}A). The forms are defined by grammatical constraints on the connections between entities; for example the ring form constrains each node to have exactly two neighbors. This approach can learn intuitive structures for a variety of domains, such as a tree for animals and a ring for the color circle. Once learned, these structures can be used to explain human inductive reasoning about novel properties of objects \citep{Kemp2009}.

Although powerful, the structural forms approach is not appropriate when the data strays from the predefined forms. Exceptions are common in real world domains. The genetic similarity of animals is captured by an evolutionary tree\footnote{Even this structure has exceptions; for example, \citet{RiveraLake2004} provide evidence that at the deepest levels ``the tree of life is actually a ring of life'' where genomes fused.}, but everyday reasoning about animals draws on factors that span divergent branches, including shared habitat, role as predator versus prey, and size. These factors cannot be perfectly explained by a single tree, and other domains are interestingly structured yet even further removed from a pristine form, such as artifacts and social networks. Since people can learn and reason in all of these domains, they must either entertain structural hypotheses without strict grammatical constraints, or engage in other types of learning.

An alternative approach is to learn implicit rather than explicit structural organization. \citet{RogersMcClelland2004} studied a neural network that learns to map animals (like canary) and relations (can) to output attributes that a canary can do (grow, move, fly, and sing). Like the structural forms approach, the neural network learns aggregate statistical structure from the observations. \citet{RogersMcClelland2004} analyzed their network's learned representation through dimensionality reduction, projecting each living thing into a low-dimensional representational space. In this space, the most salient split is between animals versus plants, with sub-divisions for mammals versus birds, trees versus flowers, etc. Although the network is not constrained or biased to learn a tree, it nonetheless learns a distributed representation for living things that exhibits hierarchical structure.

Although neural networks are powerful implicit learners, people also seem to learn and reason with explicit structural representations. \citet{Kemp2009} point out that language often carries direct structural information such as ``Indiana is next to Illinois'' or ``A dolphin is a mammal not a fish,'' observations that can be elegantly incorporated in more explicit representations. These representations can also provide scaffolding for learning higher-level concepts with direct structural interpretations, such as using a tree structure to help learn the world ``primate'' \citep{Xu2007}. Finally, it remains mysterious how implicit structure in a neural network might, in rare yet pivotal moments in science and childhood, crystallize to create explicit structural discoveries.

Here, we present a new computational approach to structure learning and discovery that incorporates some of the best features of previous probabilistic and connectionist models. Rather than selecting between discrete and mutually exclusive structural hypotheses defined by grammars, the model learns explicit structure in an unrestricted hypothesis space of all possible graphs. A key insight is that many cognitively natural structural forms are very sparse, meaning that the graphs have relatively few edges (connecting lines between the nodes). This observation is central to our new approach, called \emph{structural sparsity}, which is capable of rich structural inferences guided by a preference for sparsity. Sparsity suggests an alternative means for learning pristine structural form (Fig. \ref{fig_synthdata}A), albeit without explicitly identifying which form the structure belongs to (“ring,” “tree,” etc. as in \citet{Kemp2008}). This property of our approach has both appealing features and drawbacks, depending on how we would like to interpret it as a cognitive model. We take up these issues in detail in the discussion, at the end of the paper. Whether viewed as a strength or weakness, the lack of an explicit representation of abstract structural forms does go hand in hand with our model’s ability to explain how structure learning could proceed in domains, such as artifacts, that may not conform to any simple, cognitively natural form (Fig. \ref{fig_synthdata}B).

\begin{figure}
\centering
\includegraphics[width=6.5in]{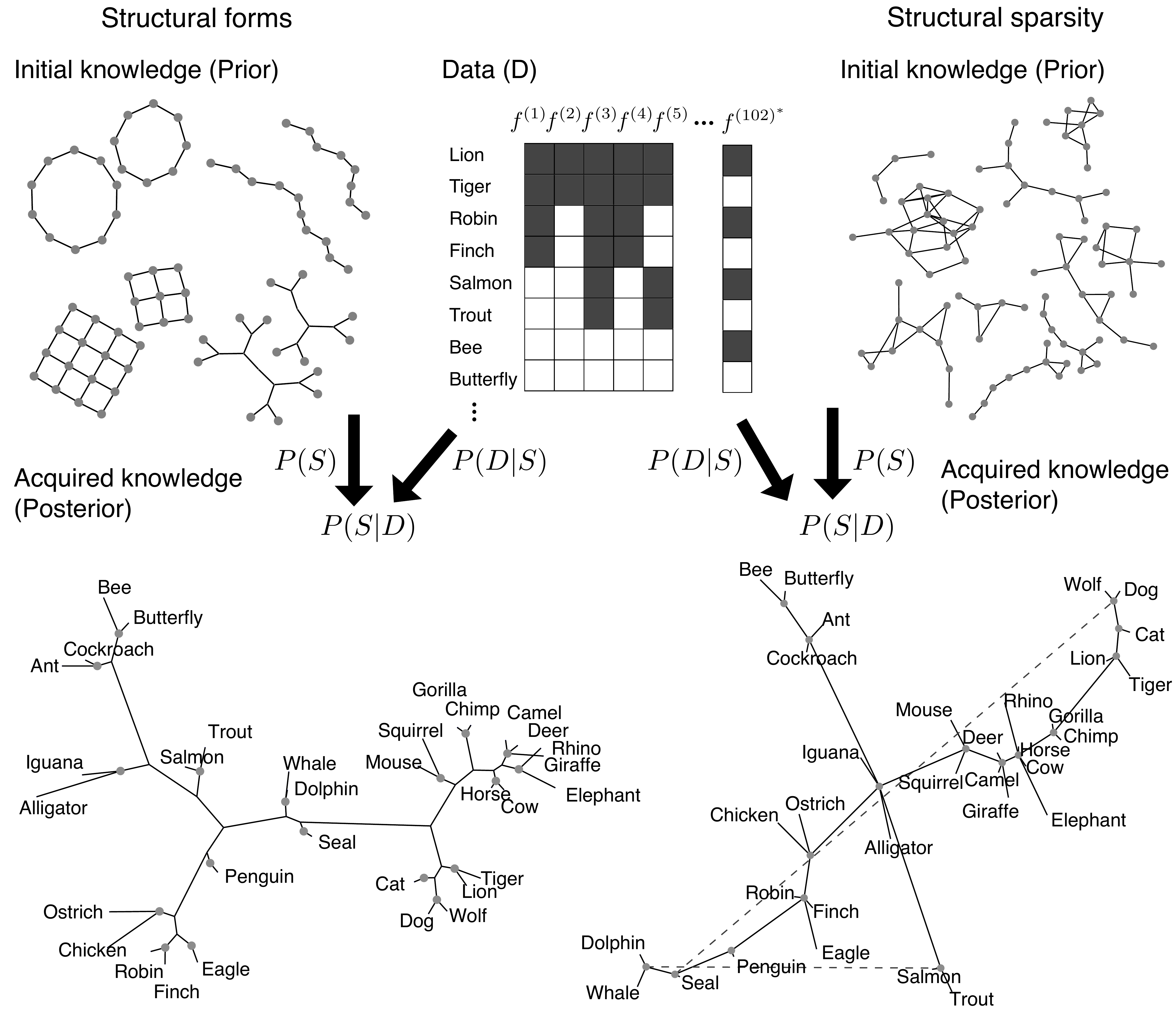}
\caption{Two models of structure discovery: (left) structural forms \citep{Kemp2008} and (right) structural sparsity (this work). Bayesian learning of structure from data $D$ involves an update from prior $P(S)$ to posterior $P(S|D)$. Given a matrix of animals and their features, the two models learned similar structures despite the substantial difference in initial knowledge (prior). In the visualizations, stronger edges are shorter, except for the dotted lines that are edges that lie outside the strongest spanning tree.
\label{fig_inductive_bias}}
\end{figure}

We also use the comparison between structural sparsity and structural forms as an opportunity to explore two different types of inductive bias and their roles in supporting learning. For any learning algorithm, it is instructive to compare the initial knowledge (before data) with the final knowledge (after data), as means of distinguishing what the model learns from what the model starts with. For the structural forms model (Fig. \ref{fig_inductive_bias} left), samples from the prior are already highly structured graphs (rings, grids, trees, etc.), suggesting that the primary role of learning is to select one of the provided forms, while simultaneously organizing the entities subject to the corresponding grammatical constraints. In contrast, for the structural sparsity model (Fig. \ref{fig_inductive_bias} right), the model's initial knowledge shows little qualitative resemblance to many of the structures it discovers, resembling some of the representational leaps in science and cognitive development \citep{Carey2009}. By comparing structural forms and structural sparsity on the same data sets -- two closely related models with different inductive biases -- we can distinguish between the cases where a strong prior over forms is needed, from other cases where a more general bias towards sparsity can lead to rich structural insight.

\section{Model}

\subsection{Structural sparsity}
Bayesian learning underlies both the structural forms and structural sparsity models, which use Bayes' rule to reason about the distribution of structural hypotheses $S$ in light of data $D$,
\begin{equation} \label{bayes}
P(S|D) \propto P(D|S)P(S).
\end{equation}
Both models use a similar likelihood $P(D|S)$ with different priors $P(S)$. The prior for the structural forms model is a mixture of discrete forms $F$ (such as rings, trees, grids, etc.), where $P(S) = \sum_F P(S|F) P(F)$. The initial knowledge $P(S)$ reflects the fact that the forms are provided explicitly and no other possibilities are entertained (Fig. \ref{fig_inductive_bias} left). The prior for structural sparsity covers a broader hypothesis space that includes all possible connectivity configurations (Fig. \ref{fig_inductive_bias} right). Sparsity is operationalized with an improper prior that decreases the score with each additional edge in a structure
\begin{equation}
P(S) \propto \exp(-\beta (\#S) ),
\end{equation}
where $\#S$ is the number of edges in a structure. In both models, objects are assigned to latent clusters, known as cluster nodes \citep{Kemp2008} (gray nodes in Fig. \ref{fig_inductive_bias}). Cluster nodes can connect to other cluster nodes in arbitrary patterns, but each object must connect to exactly one cluster node. By using cluster nodes, groups of similar objects can be elegantly represented without dense object to object connections, leading to sparser graphs that highlight the underlying topology. Besides the cluster assignments, objects do not form any other direct connections, unlike an earlier version of our model without cluster nodes \citep{Lake2010}. Given that the number of edges typically grows with the number of clusters (and clusters must be non-empty), the structural sparsity prior tends to favor both fewer edges and fewer clusters.

The term $P(D|S)$ specifies the link between the graph structure and the features of objects. The data $D = \{f\hi{1},...,f\hi{m}\}$ is a matrix where rows are objects and columns are features $f\hi{i}$. For instance, we used a data set of 33 animals and 102 features, where the first five features were ``has lungs,'' ``has a large brain,'' ``has a spinal cord,'' ``is warm-blooded,'' and ``has teeth'' (Fig. \ref{fig_inductive_bias} top). The distribution $P(D|S)=\prod_{i=1}^m P(f\hi{i} | S)$ treats features as independent draws from a distribution that prefers smoothness, meaning objects tend to share similar feature values with their neighbors in the graph. For example, all the features shown in Fig. \ref{fig_inductive_bias} are relatively smooth (except for a hypothetical $f\hi{102^{*}}$ which is highly non-smooth) over the graphs that represent animals (Fig. \ref{fig_inductive_bias} bottom). An edge between node $i$ and $j$ is parameterized by a continuous value $s_{ij}>0$ that determines its strength, and absent edges have $s_{ij}=0$. The graph as a whole specifies the covariance matrix of a multivariate Gaussian distribution with one dimension per node (Appendix \ref{appendix_gen_model}). Although $P(D|S)$ is best suited for continuous features, binary data can be treated as continuous \citep{Kemp2008}. Data in the form of a similarity matrix can also be accommodated by treating the matrix as the covariance matrix of a multivariate Gaussian, and then sampling an input data matrix $D$ from that Gaussian (we used $m=2000$ features).
 
As in the structural forms model, we are interested in the single best structure given the data. For structural sparsity, taking the log of the posterior probability (Eq. \ref{bayes}) leads to the optimization problem
\begin{equation} \label{score}
\amax{S} \sum_{i=1}^m \log P(f\hi{i} | S) - \beta(\#S),
\end{equation}
which emphasizes the dual objectives of capturing regularities in the features (left term) while promoting the sparsity of the structure (right term). The parameter $\beta$, which is the only free parameter in the model, controls the trade-off between the two objectives. The value $\beta=6$ was used for most of the simulations in this paper. For large data sets with many objects, a larger value leads to more interpretable structural discoveries ($\beta=18$ was used in two cases). Beyond these guidelines, we do not know if there is a completely general procedure for setting $\beta$ to facilitate interpretable structural discoveries, and this is a limitation of our work currently.

\subsection{Search algorithm}
Searching for the maximum of this score (Eq. \ref{score}) is a special case of learning the structure of an undirected graphical model, a generally intractable problem that requires heuristic algorithms not guaranteed to reach the global optimum \citep{Koller2009}. Here we describe the heuristic search algorithm developed for solving this problem. At its core, structural sparsity is a proposal for the computational level problem \citep{Marr1982} of structure discovery, and thus the particular search algorithm we used was not intended to be cognitively plausible.

Our search algorithm consists of an outer-routine that looks for the best partition of objects into cluster nodes, as well as a sub-routine that evaluates a candidate partition by optimizing for the best set of edges. The outer-routine tries local proposals for splitting and merging clusters and greedily picks the best move at each step \citep{Kemp2008}. Although searching over partitions can be challenging, it can be sped up greatly by a preprocessing stage that clusters objects several times, with different numbers of clusters, and picks the best scoring option for initialization.

The sub-routine has the challenging task of estimating the sparse connectivity between clusters, necessary for evaluating partition quality during both initialization and local search. Directly searching the combinatorial space of sparsity patterns is intractable, but there are well-known heuristics such as $\ell_1$ that push parameters to zero ($s_{ij}=0$) and consequently propose a sparsity pattern \citep{Tibshirani1996}. The $\ell_1$ relaxation penalizes the sum rather than the number of parameters $\#S \approx \lambda \sum_{ij} |s_{ij}|$ for some constant $\lambda$. Learning a Gaussian graph with $\ell_1$ penalization from a complete data matrix is a convex optimization problem that can be solved efficiently \citep{Banerjee2008,Lake2010}. Although our connectivity estimation problem is not convex due to the latent nodes and missing data, efficient solvers for the complete data case are utilized in a Structural Expectation-Maximization algorithm guaranteed to improve the primary objective in Eq. \ref{score} with each iteration \citep{Friedman1997}.

The Appendix (Section \ref{appendix}) contains a more detailed description of the entire search algorithm. Although search can be slow, it considers the full range structures, unlike the implementation of structural forms that runs a separate search for each form \citep{Kemp2008}. Code for running the model is available online.\footnote{\url{https://github.com/brendenlake/structural-sparsity}}

\section{Discovering structure in synthetic data}
The algorithm was evaluated on its ability to recover the true topology of synthetic data. Given that objects can cluster in a variety of patterns atop these forms, the challenge of jointly finding the right cluster partition and the right sparsity pattern results in a very difficult combinatorial search problem. Structural sparsity was tested on both its ability to recover structural forms (Fig. \ref{fig_synthdata}A) as well as novel structures that share interesting features with the forms, but not the pristine qualities necessary for identification with the \citet{Kemp2008} approach (Fig. \ref{fig_synthdata}B). These new structures include a tree embedded in a ring (i), a non-grid planar graph (ii), a ring of trees (iii), and multiple disconnected forms including disjoint chains (iv; Fig. \ref{fig_synthdata}B).

The synthetic experiments were structured as follows. For each candidate synthetic structure, 1000 features were generated from the ground truth graph and provided to the model. The sparsity free parameter was set to $\beta=6$. Given that the algorithm is non-deterministic, the search process was repeated 10 times for each synthetic data set. Each run of the search algorithm is compared to the ground truth.

\begin{figure}
\centering
\includegraphics[width=6.5in]{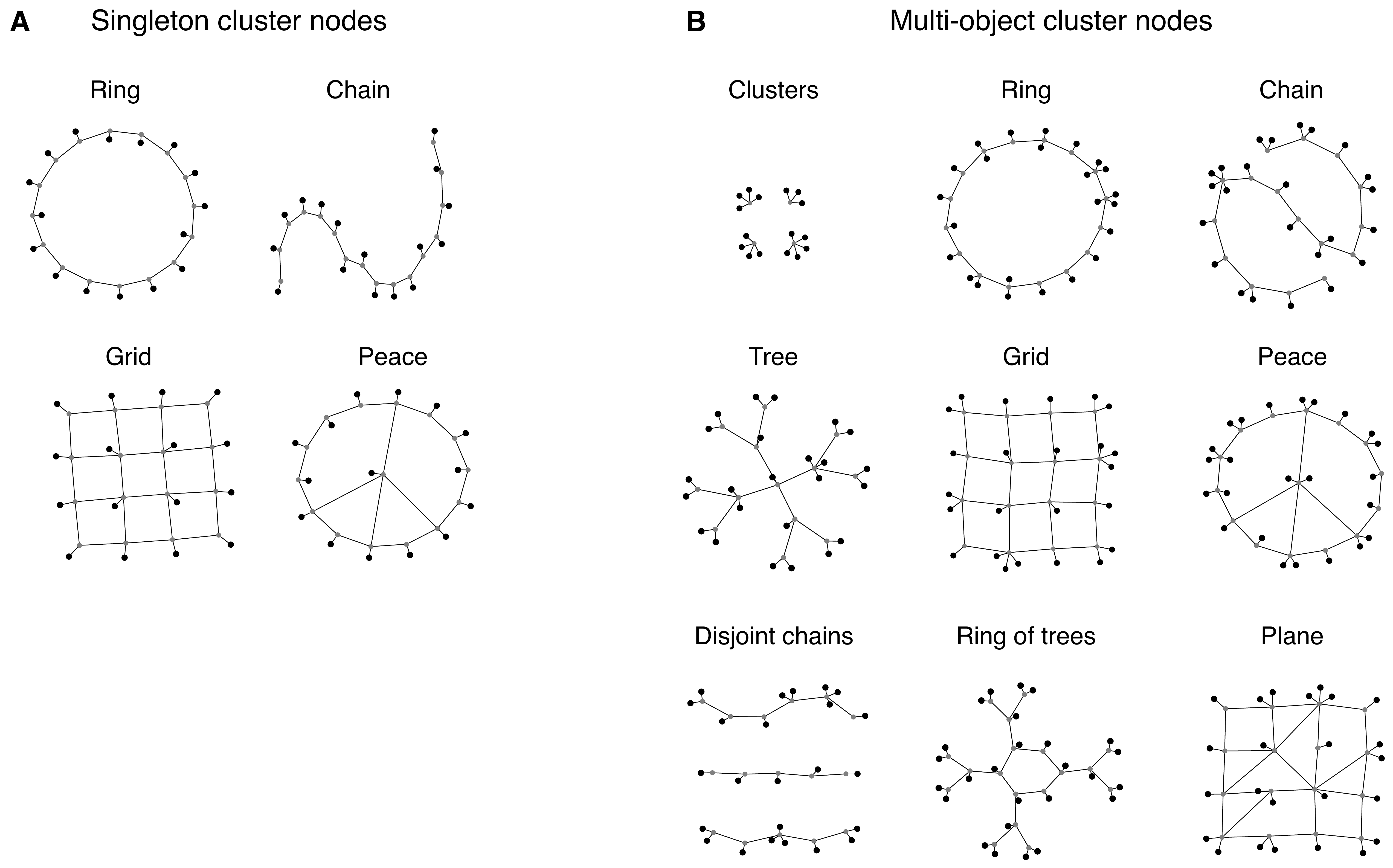}
\caption{Synthetic data sets were generated from structures with one object per cluster node (A) and arbitrary clusters of objects on the underlying structure (B).}
\label{synthetic}
\end{figure}

The structures in the first set of experiments had exactly one object per cluster node (Fig. \ref{synthetic}A). This clustering pattern is easier for the search algorithm to find compared to an arbitrary clustering, in part because it is automatically examined as a candidate pattern for initialization.
Thus, this helps to isolate the challenge of finding the right sparsity pattern from the problem of finding the right clustering. The results of search are shown Table \ref{synth_single}. In each case, algorithm recovers a structure that scores (Eq. \ref{score}) at least as well as the original structure (Table \ref{synth_single} Column 1, score match).\footnote{Structures are counted as correct if they are within 4 log points of the ground truth, so that an extra edge will be counted as an error (-6 log points).} For all cases except the chain, the highest scoring structure recovered the right partition of objects into cluster nodes (Table \ref{synth_single} Column 2, partition match), and furthermore the structure matched the adjacency matrix up to a permutation of the cluster node indices (Table \ref{synth_single} Column 3, exact match). There is straightforward reason why the chain structure was not recovered exactly. To increase sparsity, the model tended to combine the last two objects on each end of the chain into a single cluster, using a weaker connection to the object at the end of the chain. This configuration retains the correct relations while increasing sparsity, and thus the learned structure scores higher than the original graph.\footnote{This reduction also occurred in the original structural forms model, in order to use fewer cluster nodes.}

The structures in the second set of experiments had richer clustering patterns on the same graphs, as well as others like clusters, a tree, a peace sign, disjoint chains, a ring of trees, and a plane (Fig. \ref{synthetic}B). The results are shown in Table \ref{synth_multi}. The clusters, ring, tree, and ring of trees were perfectly recovered for each run. Like in the singleton case, the algorithm always conserved edges on the ends of the chain structure and disjoint chains structure. Mistakes occurred on 1 run of the grid, 2 runs of the peace graph, 1 run of the plane, and all of these were unnecessary cluster node splits that did not alter the underlying topology. For several runs of the plane structure, the model recovered an alternative graph that differed in the location of one edge, and it also scored roughly as well as the ground truth. On the whole, our search algorithm was highly successful at recovering ground truth structure from synthetic data.

\begin{table}[h]
\centering
$\begin{array}{lccc}
\hline & \text{Score match} & \text{Partition match} & \text{Exact match} \\ \hline
\text{Ring}    &10 & 10 &  10 \\
\text{Chain}  &10 & 0   &   0  \\
\text{Grid}     &10 & 10 &  10 \\
\text{Peace} &10 & 10 &  10 \\ \hline
\end{array}$ 
\caption{Synthetic experiments with singleton cluster nodes (Fig. \ref{synthetic}A).
Entries show the number of successes out of 10 runs of the search algorithm.}
\label{synth_single}
\end{table}

\begin{table}[h]
\centering
$\begin{array}{lccc}
\hline  & \text{Score match} & \text{Partition match} & \text{Exact match} \\ \hline
\text{Clusters}   &10 & 10 &  10 \\
\text{Ring}  & 10 &   10  &  10 \\
\text{Chain}  &10  &   0  &   0 \\
\text{Tree}   &10 & 10 &  10 \\
\text{Grid}   & 9   & 9  & 9 \\
\text{Peace}  & 8  & 8 &  8 \\ 
\text{Disjoint} & 10 & 0 & 0 \\
\text{\ chains} \\
\text{Ring} & 10 & 10 & 10 \\
\text{\  of trees} \\
\text{Plane} & 9 & 9 & 3 \\
\hline \end{array}$
\caption{Synthetic experiments with multi-object cluster nodes (Fig. \ref{synthetic}B).
Entries show the number of successes out of 10 runs of the search algorithm.}
\label{synth_multi}
\end{table}

\section{Discovering structure in empirical data}
Structural sparsity was applied to six empirical data sets. To facilitate comparison with the structural forms approach, four data sets from \citet{Kemp2008} were included. The search algorithm was run ten times for each data set and the highest scoring structure is displayed. The sparsity prior was fixed at $\beta = 6$. In each case, the output of the structural forms model is displayed for comparison. The data sets were rescaled following the procedure of \citet{Kemp2008}.\footnote{Each data matrix $D$ ($n_x$ objects by $m$ features) was linearly transformed so the mean across the entire matrix is 0 and the largest element in $\frac{1}{m}DD^T$ is 1. If there are missing values, the features with the most common pattern of missing objects are grouped, and the transform is computed for this sub-matrix and applied to the full matrix.}

\begin{figure}
\centering
\includegraphics[width=6.5in]{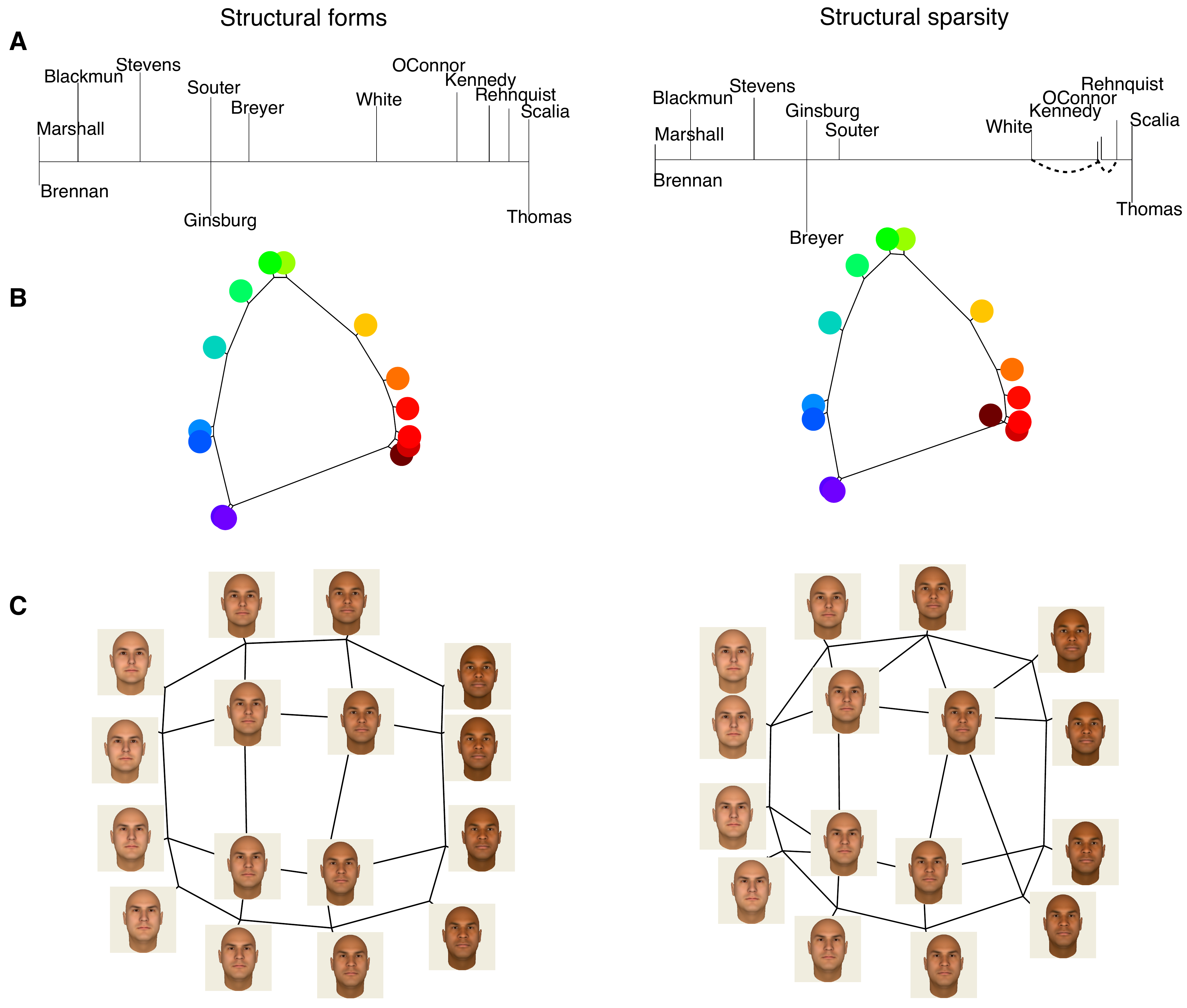}
\caption{Structures learned by structural forms (left) and sparsity (right) for a vote matrix from 13 Supreme Court justices (A), similarity judgments between spectral colors (B), and Euclidean distance between images of faces (C). To visualize the spatial structure, node locations were chosen to be the principal eigenvectors of the implied covariance matrices \citep[see][]{Lawrence2011,Lawrence2012}. Only one dimension is shown in (A), and dotted lines are edges that do not follow a rigid linear order.}
\label{fig_main_set}
\end{figure}

\subsection{Animals}
The first data contains a set of 33 animals with 102 features related to biology, anatomy, and habitat \citep[collected by ][]{Kemp2008}. Biological features included those like ``has lungs'' and ``is warm-blooded,'' anatomical features included ``has fins'' and ``has a long neck,'' and perceptual features included ``is black.'' Habitat features often highlight non-biological similarity, like the fact that fish and the aquatic mammals ``live in the ocean.''

The learned structures are shown in Fig. \ref{fig_inductive_bias}. Although structural forms learned a tree, structural sparsity learned a graph with a tree-based backbone that primarily reflects biology, with branches for the insects, fish, birds, and mammals. Structural sparsity can also capture relationships that reach across the tree. The aquatic mammals are situated with the fish, penguin, and other mammals, representing both biology, shared habitat, and visual similarity. The tree-based structural form better reflects evolutionary branching, where objects must lie at the leaf nodes. Trees such as this are not considered by structural sparsity, since allowing empty cluster nodes would make search much more difficult. But these form-based latent trees are not necessarily highly sparse, and here structural sparsity found about half as many non-attachment edges.

\subsection{Judges}
The second data set included 13 Supreme Court justices and their votes on 1,596 cases, collected and preprocessed by \citet{Kemp2008} to include 1596 cases under Chief Justice Rehnquist. Due to non-participation and the fact that at most 9 justices serve at a time, there were many missing values. We integrated over the missing values while computing the model score.

The learned structures are shown in Fig. \ref{fig_main_set}A. Both models organized the justices along a spectrum from liberal (Marshall and Brennan) to conservative (Scalia and Thomas). Although structural forms is constrained to learn exactly a chain, structural sparsity learned a similar chain with a few additional relationships between the conservatives (two additional dotted edges and one gap).

\subsection{Colors}
The next data set includes perceived similarity ratings between 14 spectral colors. Originally collected by \citet{Ekman1954}, the similarity matrix was published in \citet{Shepard1980} and used in \citet{Kemp2008}, while assuming a value of 1 along the diagonal. Provided with this data, the structural sparsity model discovered the color circle first described by Newton (Fig. \ref{fig_main_set}B). Although the ring topology was pre-specified in the forms model, it is an emergent consequence of the data combined with structural sparsity.

\subsection{Faces}
Another data set is based on pixel similarity between images of faces that vary along a masculinity and a race dimension. This similarity data set was created by \citet{Kemp2008}. The faces vary in two dimensions, race and gender, with four values along each. The similarity matrix between faces is based on the Euclidean distance in pixel space, with 1 along the diagonal. Structural sparsity recovered these two dimensions, although it deviates from a strict grid (Fig. \ref{fig_main_set}C).

\begin{figure}
\centering
\includegraphics[width=6.5in]{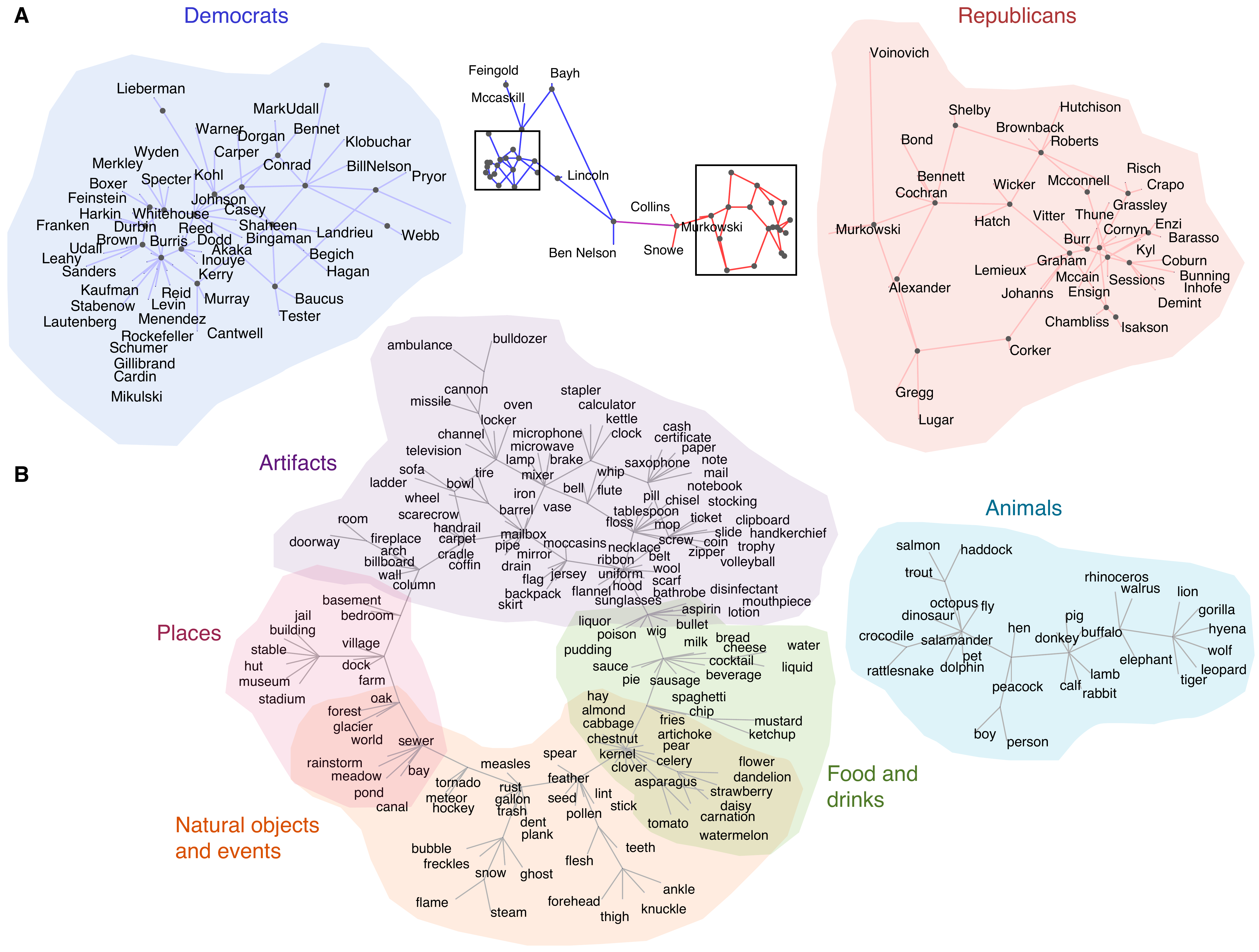}
\caption{Sparse structures learned from two data sets: US Senators and their votes in the 111th congress (A), and 200 objects and their features with broad semantic coverage (B). In (A), shorter edges are stronger, but in (B) all edges are shown at approximately the same length for visual clarity. Some coherent sub-regions are shaded and labeled for easier interpretation; these are not labels provided to the unsupervised learning algorithm.}
\label{fig_large_set}
\end{figure}

\subsection{Senate}
Structural sparsity can also be applied to domains that have a rich organization but do not fit a cognitively natural structural form. We used a data set of Senators and their voting records was for the 111th United States Congress, from January 2009 to January 2011.\footnote{The votes were retrieved from \url{http://www.voteview.com/house111.htm}. The various yes votes (1,2,3) and no votes (4,5,6) were collapsed to yes (1) or no (0).} Voting present or not voting was treated as missing data. Only senators that were present for at least half the votes were included, resulting in 98 Senators and 696 votes. Due to the size of this data set (and the following), the sparsity parameter was increased to $\beta = 18$ to aid interpretability. To maintain tractability, search was also modified to be less thorough, as described in Appendix \ref{appendix_search}.

Given the voting records, the model recovered the central divide between Democrats and Republicans, bridged only by a connection between moderate Democrat Ben Nelson and Republicans Olympia Snowe and Susan Collins (Fig. \ref{fig_large_set}A). Although structural forms might capture the party distinction along a line or with clusters, structural sparsity suggests that the complex within-party dynamics are not naturally represented by any one of the pristine forms.

\subsection{Common objects}
Structural sparsity was also applied to 200 common objects with broad semantic coverage across artifacts and living things \citep{Palatucci2009}. Each object was rated according to 218 properties, including questions like ``Is it manmade?'', ``Can you hold it?'', and ``Does it have feet?'' Answers were on a 5 point scale from ``definitely no'' to ``definitely yes'' conducted on Amazon Turk. Using a conservative threshold for indicating the presence of a feature, the data set was converted to binary form by coding only the most confident response as 1 and the rest as 0. The original data set contained 1000 objects, and a subset of 200 objects was randomly selected for tractability.

Given the object and feature data, structural sparsity learns a complex structure with sub-regions for artifacts, animals, food and drinks, natural objects and events, and places (Fig. \ref{fig_large_set}B). Edges instantiate a notion of aggregate semantic relatedness, resembling classical proposals for semantic networks \citep{Collins1975}. Although the data set is prohibitively large for comparison with the structural forms algorithm, it would likely miss much of the fine-grained structure in an attempt to fit the objects into one of the pre-conceived forms.

\section{Property induction} \label{sec_prop_induction}
Structural sparsity, in addition to revealing organizing form, can be used to predict how people reason about novel properties. First, structures were learned to represent two data sets, one for mammals and one for cities. Second, the learned structures were used to make predictions about how people reason in a property induction task.

Inductive questions looked like the following: given that a new biological property is true of dolphins, squirrels, and chimpanzees, how likely is it true of all mammals? \citep{Rips1975, Osherson1990, Heit1998, Heit2000, Kemp2009}. Earlier work using the structural forms approach found an interesting double dissociation \citep{Kemp2009}: people's ratings of argument strength for mammals were better predicted by a tree than a continuous 2D space, while analogous ratings regarding geographical properties of cities were better predicted by a 2D space than a tree. Here, we investigated whether structural sparsity can capture both patterns of reasoning without requiring special purpose structural forms.

\subsection{Data sets for structure learning}
The mammals data set consists of 50 mammals and 85 features (properties), primarily collected by \citet{Osherson1991} and extended in \citet{Kemp2009}. The features contain biological, anatomical, behavioral, and habitat properties (``is gray'', ``has tough skin'', ``big'', ``swims'', ``lives in water''). Features were collected by asking participants to rate the strength of association between each animal and feature, starting at zero and imposing no upper bound. Ratings were scaled between 0 and 100 and then averaged across participants.

The cities data set is a similarity matrix of 9 US cities collected by \citet{Kemp2009}. Participants were asked to draw the locations of the 9 cities on a piece of paper, ensuring that the relative distances were as accurate as possible. The distances for each participant were scaled so that the largest distance was 1, and then each value was subtracted from 1 to create a measure of spatial similarity. The matrix was averaged across participants and then provided as input to the models.

\subsection{Models for structure learning}
Four alternative models were trained on the two base data sets: structural sparsity, a tree structural form, a 2D spatial model, and the raw covariance matrix. For each of the models, the product of learning is a multivariate Gaussian distribution with one dimension per object. Each multivariate Gaussian parameterizes a joint distribution on new object properties, which can be used to make Bayesian predictions regarding the strength of inductive arguments. All inductive arguments use new (generic) properties that were not features in the data set used for structure learning.

\begin{itemize}

\item \textbf{Structural sparsity}. Structural sparsity represents a Gaussian by a sparse graph (Section \ref{appendix_gen_model}). The structural sparsity model has one free parameter $\beta$, which controls the degree of sparsity in the learned structure. The model was run using a range of different sparsity values, and the fit to the human property induction judgments were reported for each value. For each value of $\beta$, the search algorithm was run twice for mammals and cities, and the highest scoring structure (based on the score in Equation \ref{score}, not correlation with human participants) was chosen to represent that value of sparsity. Examples of learned structures for mammals are shown in Fig. \ref{mammals_graphs}A \& D and for cities in Fig. \ref{fig_induct}B-i.

\item \textbf{Tree-based structural form}. For both data sets, we assumed a tree-based form and searched for the best tree using the algorithm from \citet{Kemp2009}. The learned structures are shown in Fig. \ref{mammals_graphs}B and Fig. \ref{fig_induct}B-ii. Given that the transformation from a graph to a Gaussian differs in some minor ways between structural forms and structural sparsity, we relearned the edge strengths for the tree using maximum likelihood and the graph formalism from structural sparsity. However, re-learning the edge weights did little to change the predictive performance on the property induction tasks.

\item \textbf{2D spatial form}. For both data sets, we assumed a 2D spatial form and searched for the best spatial representation using the algorithm from \citet{Kemp2009} to compare with the graph-based methods. The learned representational spaces shown in Fig. \ref{mammals_graphs}C and Fig. \ref{fig_induct}B-iii. This model represents a covariance matrix $\Sigma$ using distances in a 2D space
\begin{equation}
\Sigma_{ij} = \frac{1}{2\pi} \exp(-\frac{1}{\sigma} || x_i -x_j||),
\end{equation} 
where $x_i$ is the 2D location of object $i$ and $||x_i - x_j||$ is the Euclidean distance between two objects.

\item \textbf{Raw covariance}. The raw covariance model was either identical to the similarity matrix or $\Sigma = \frac{1}{m}DD^{T}$ where $D$ is the rescaled feature matrix.

\end{itemize}

\subsection{Data sets for property induction}
The property induction data concerning mammals, including the Osherson horse and Osherson mammals tasks, was reported in \citet{Osherson1990}. Judgments concerned 10 species: horse, cow, chimp, gorilla, mouse, squirrel, dolphin, seal, and rhino. Participants were shown arguments of the form ``Cows and chimps require biotin for hemoglobin synthesis. Therefore, horses require biotin for hemoglobin synthesis." The Osherson horse set contains 36 two-premise arguments with the conclusion ``horse,'' and the mammals set contains 45 three-premise arguments with the conclusion ``all mammals." Participants ranked each set of arguments in increasing strength by sorting cards.

The induction task concerning cities was conducted by \citet{Kemp2009}. Participants were presented a scenario where Native American artifacts can be found under most large cities, and some kinds of artifacts are found under just one city while others are under a handful of cities. An example inductive argument is: ``Artifacts of type X are found under Seattle and Boston. Therefore, artifacts of type X are found under Minneapolis.'' There were 28 two-premise arguments with Minneapolis as the conclusion, 28 with Houston as the conclusion, and 30 three-premise arguments with ``all large American cities'' as the conclusion. These arguments were ranked for strength by sorting cards, in a method intended to mimic the Osherson tasks.

\subsection{Bayesian property induction}
We used a Bayesian model of property induction that takes a structured representation (sparse graph, tree, space, or raw covariance) and uses it to make predictions in the property induction tasks \citep{Kemp2009} \citep[see also][]{Heit1998}. We refer the reader to \citet{Kemp2009} for a fuller treatment, but the basics are described here. 

Inductive arguments involve questions of the form: Objects $X$ have property $f$, therefore, how likely is it that objects $Y$ have property $f$? The first set of objects $X$ are the premise categories, and the second set $Y$ is the conclusion category. The argument concerns a new feature $f$ which is observed on the set $f_X$ to be equal to the label $l_X$ (here $l_X = [1,\ 1,\ 1]$ since all the premise objects have this property). With this notation, the strength of an inductive argument can be modeled as the probability the property is true for the conclusion $f_Y = 1$ given the premise labels $f_X = l_X$, or
\begin{equation}
P(f_Y = 1 | f_X = l_X) = \sum_{f: f_Y = 1} P(f | f_X = l_X),
\end{equation}
where the posterior $P(f | f_X = l_X)$ is the distribution on full instantiations of the binary feature vector $f$, given the labeled premises. This distribution can be computed by Bayes' rule
\begin{equation}
P(f | f_X = l_X) = \frac{ P(f_X = l_X | f) P(f)}{\sum_f P(f_X = l_X | f) P(f)}.
\end{equation}
The likelihood $P(f_X = l_X | f) \propto 1$ if the label is consistent with the feature, and otherwise it is 0. The prior distribution $P(f)$ is instantiated by each of the different models. It follows that the strength of the inductive argument can be computed as 
\begin{equation}
P(f_Y = 1 |f_X = l_X) = \frac{\sum_{f: f_Y = 1, f_X = l_X} P(f)}{\sum_{f: f_X = l_X} P(f)}.
\end{equation}
Intuitively, this is the weighted fraction of features consistent with the premises and conclusion, compared to those consistent with just the premises. Each feature is weighted by its prior probability $P(f)$. Given that each model specifies a Gaussian distribution on features rather than a distribution on binary features, this prior $P(f)$ was approximated by a large number ($10^6$) of continuous Gaussian samples, thresholded at the mean (0) to create binary features.

\subsection{Results}

\begin{figure}
\centering
\includegraphics[width=6.5in]{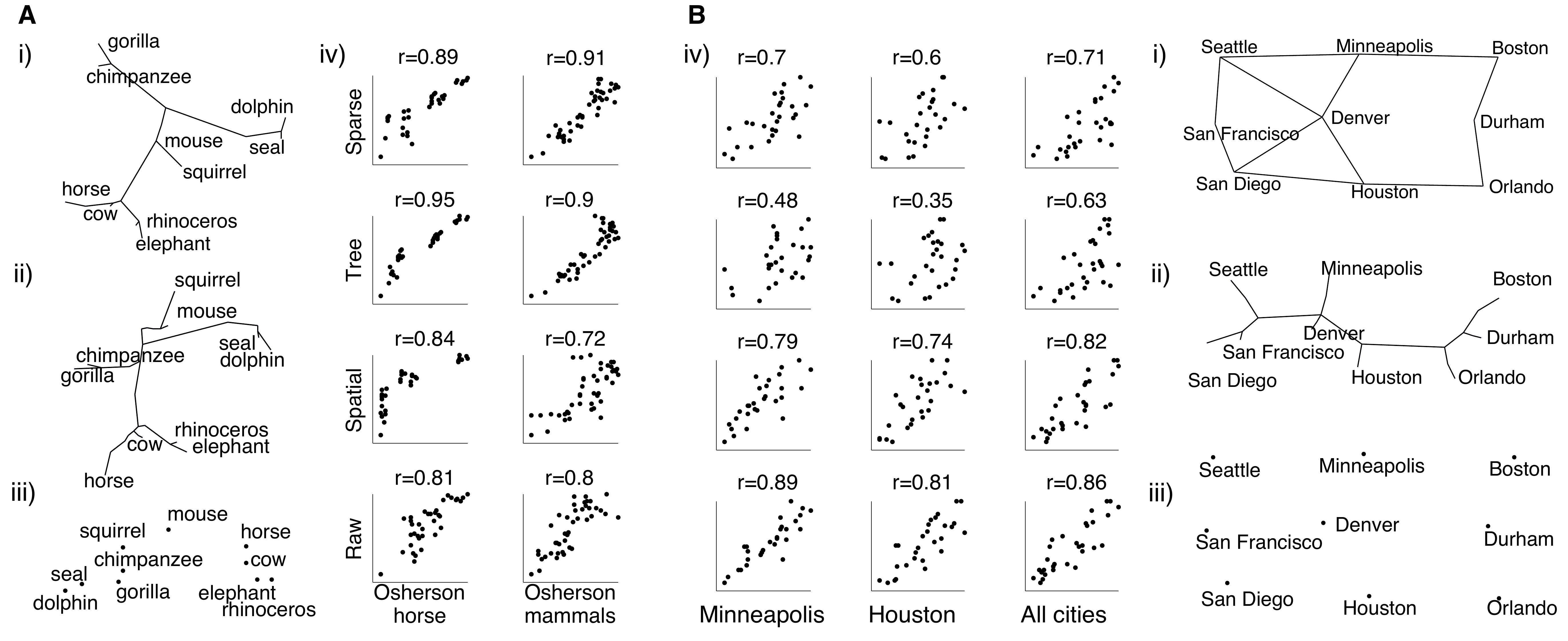}
\caption{Comparing models and human responses regarding property induction for mammals (A) and cities (B). A dot in (iv)  is an inductive argument; for instance, the bottom-left dot for the ``Osherson horse'' data set is the argument ``dolphins and seals have this property, therefore horses do.'' Each dot represents a different set of premises, followed by the conclusion category labeled on the x-axis (horse, all mammals, Minneapolis, Houston, or all cities). Argument strength for the models (x-axis) is plotted against mean rank of strength across participants (y-axis), with the correlation coefficient $r$ shown above. Predictions were compared for different types of structures, including those learned with structural sparsity (i), trees (ii), and 2D spaces (iii). The arguments about mammals mention only the 10 mammals shown in (A), although predictions were made by using the full structures learned for 50 mammals. Here only the subtrees (or space) that contains the pairwise paths between these 10 mammals are shown.}
\label{fig_induct}
\end{figure}

The models were evaluated on their ability to predict argument strength as evaluated by participants. Predictive performance was evaluated as the Pearson correlation between people's rankings and the models' evaluation of argument strength. The structural sparsity predictions using $\beta=8$ for mammals and $\beta=4$ for cities are highlighted in Fig. \ref{fig_induct} and compared with the other models. For mammals, structural sparsity predicted human ratings about as well as the tree form (Fig. \ref{fig_induct}A-iv). For cities, structural sparsity fit much better than a tree and somewhat worse than a 2D space (Fig. \ref{fig_induct}B-iv). The raw covariance model fit particularly well for cities, since the similarity data was already based on 2D drawings, but did not fit as well for the mammals data sets.

\begin{figure}
\centering
\includegraphics[width=6.5in]{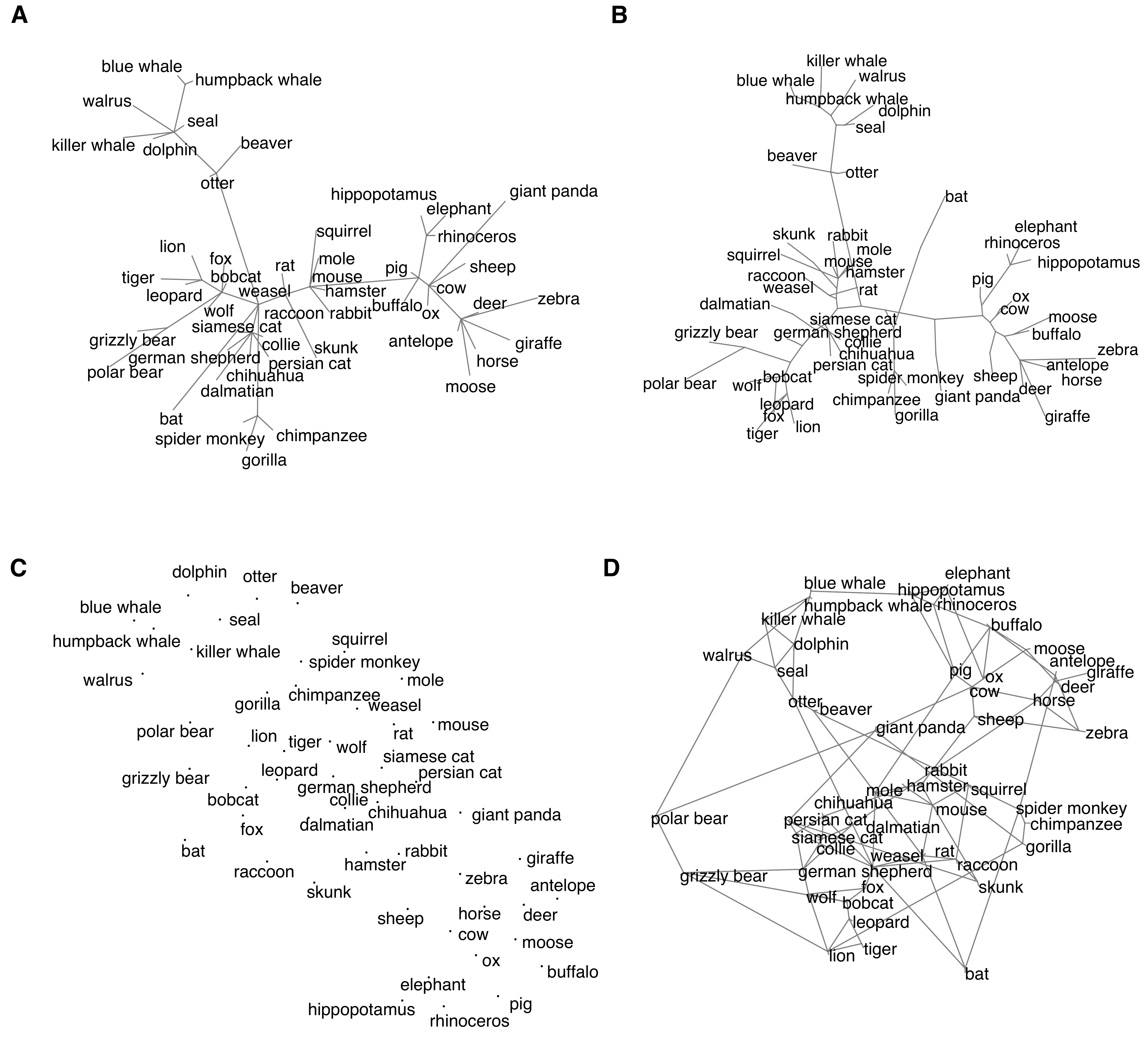}
\caption{Learned structures to represent the mammals data set. Structural sparsity with $\beta=8$ (A), tree-based structural form (B), 2D spatial structural form (C), and structural sparsity with $\beta=1$ (D). Stronger edges are displayed as shorter.}
\label{mammals_graphs}
\end{figure}

It is noteworthy that while structural forms assumes different forms for the two domains, structural sparsity provides comparable fits with a more general inductive bias, suggesting that domain-tailored forms may not be necessary to explain these judgments. Structural sparsity fit well across the range of values of its sparsity parameter (Table \ref{induct_beta}), and that these different sparsity settings result in qualitatively different structures for the mammals. For low sparsity ($\beta=1$, Fig. \ref{mammals_graphs}D), the model groups similar mammals together, but it is otherwise difficult to interpret. For higher sparsity ($\beta=8$, Fig. \ref{mammals_graphs}A), the structure closely mimics the tree-based structural form (Fig. \ref{mammals_graphs}B). Structural sparsity can predict the human judgments using either structure, indicating that a pristine tree is sufficient, but not necessary, to explain the human judgments.\footnote{Similar results were obtained by learning sparse graphs without latent cluster variables in \citet{Lake2010}.}

\begin{table}
$\begin{array}{lccccc} 
\hline \beta & \text{Osherson} & \text{Osherson} &  \text{Minneapolis} &  \text{Houston} & \text{All} \\ 
& \text{horse}        & \text{mammals}  &  & & \text{cities} \\ \hline
1 & 0. 94 &  0.87 &  0.70 &   0.59 &   0.71\\
2 &  0.95 &  0.86  &  0.70  &  0.58  &  0.71\\
4  &  0.93  &  0.84 &   0.70 &   0.61 &   0.71\\
6 & 0.91  &  0.89 &   0.69  &  0.51 &   0.68\\
8  &  0.89  &  0.91  &  0.69 &   0.51 &   0.68\\ \hline
\end{array}$
\caption{Correlations between how people and the structural sparsity model judge inductive strength, for several tasks concerning mammals and cities. Rows of the table indicate different values of the sparsity parameter $\beta$.}
\label{induct_beta}
\end{table}

\section{General Discussion}

As illustrated across a range of data sets, structural sparsity can discover qualitatively different organizing structures. Rather than predefining a set of forms from which it selects, the model considers a continuum of hypotheses with a simplicity bias towards sparsity. Despite this weaker inductive bias, the model can nevertheless recover distinct structures such as trees, rings, and chains for tightly organized domains. Unlike the structural forms model \citep{Kemp2008}, structural sparsity can learn exceptions to forms which often carry important semantic meaning. If a given domain is more loosely organized, such as artifacts, the model learns a complex organization that does not fit an easily recognizable form. Once a domain structure is learned, it can also be used to predict the extension of novel properties. Where previous work showed a double dissociation between a tree and a 2D form for matching people's inferences about animals and cities \citep{Kemp2009}, structural sparsity can learn different types of representations that support different patterns of prediction.

Structural sparsity is related to machine learning approaches for dimensionality reduction and feature prediction that typically operate in more restricted settings. Some algorithms for non-linear dimensionality reduction can be interpreted as using the same graphs and likelihood model as structural sparsity \citep{Lawrence2011,Lawrence2012}. But these algorithms do not have latent clusters and commonly use a fixed graph created by adjoining nearby objects. There are also efficient algorithms for learning sparse networks or sparse Gaussian graphical models, either without latent variables \citep{Hutchinson1989,SchvaneveldtDurso1989,Banerjee2008,Lake2010} or without fully recovering a latent topology \citep{Chandrasekaran2009}. Other algorithms learn trees with latent variables \citep{Choi2010,Choi2011} or jointly learn two graphs -- one for objects and one for features -- although without latent variables \citep{Kalaitzis2013}. Although these algorithms are computationally efficient and successful in the settings they consider, they do not attempt to explain how people can learn different types of organizing structure for different domains. Structural forms showed how the type of structure can be learned by discrete selection, and here we showed that different forms can arise through a general bias towards sparsity.

Structural sparsity also brings a new perspective to an old debate in cognitive science between symbolic \citep{Griffiths2010a} versus emergent \citep{McClelland2010} approaches to knowledge representation. The symbolic tradition uses classic knowledge structures including graphs, grammars, and logic, viewing these representations as the most natural route towards the richness of thought \citep{Griffiths2010a,Tenenbaum2011}. The competing emergent tradition views these structures as epiphenomena: they are approximate characterizations that do not play an active cognitive role. Instead, cognition emerges as the cooperant consequence of simpler processes, often operating over vector spaces and distributed representations \citep{McClelland2010,McClelland2010a}. This debate has been particularly lively with regards to conceptual organization, the domain studied here. The structural forms model \citep{Kemp2008} has been criticized by the emergent camp for lacking the necessary flexibility for many real domains, which often stray from pristine forms \citep{McClelland2010}. The importance of flexibility has motivated emergent alternatives, such as a connectionist network that maps animals and relations on the input side to attributes on the output side \citep{RogersMcClelland2004}. As this model learns, an implicit tree structure emerges in its distributed representations. But those favoring explicit structure have pointed to difficulties: it becomes hard to incorporate data with direct structural implications like ``A dolphin is not a fish although it looks like one'' \citep{Kemp2009}, and latent objects in the structure support the acquisition of superordinate classes such as ``primate'' or ``mammal''  \citep{Xu2007}. Structural sparsity shows how these seemingly incompatible desiderata could be satisfied within a single approach, and how rich and flexible structure can emerge from a preference for sparsity.

Although the structural sparsity model was formulated at Marr's computational level \citep{Marr1982}, any complete understanding must also extend to the algorithmic and implementation levels. Explaining how this model, and other models at the computational level, could be approximated by neural circuits is an important challenge that we do not address in this work. Nevertheless, we find it intriguing that sparsity is a general principle with broad application in both neuroscience and cognitive science, whether used to describe synaptic connectivity \citep{Kalisman2005,Fares2009}, neural activation \citep{Olshausen1996}, or environmental structure \citep{Oaksford1994,Navarro2011}. We think it is worth exploring these connections with the aim of building bridges across Marr's levels.

A final puzzle is to understand how a more explicit notion of structural form might arise. For some domains, there may be no suitable form to characterize the learned outcome of structural sparsity (e.g., Fig. \ref{fig_large_set}B), but for other domains, the learned structure can approximate the pristine output of a graph grammar. As the model currently stands, form is not explicitly identified with a label (``ring,'' ``tree,'' etc.), although such a label might be necessary for genuine ``conceptual change'' \citep{Carey2009}. An explicit structural form may also serve as a basis for analogy between domains, as a hypothesis about causal structure within a domain, or as a vehicle for easier communication of larger domain structure (``left'' or ``right'' in a chain and ``clockwise'' or ``counterclockwise'' in a ring). Somehow the mind can grasp structure at this level of intuitive description, without placing the bounds of learning around a small set of forms. 

How can both explicit form and flexibility be captured in a single computational approach, and can we extend the work developed here in this direction? One possibility is to extend the approach of \citet{Kemp2008} to achieve greater flexibility by allowing exceptions to the grammar-based forms. As an alternative to top-down search, in the spirit of the work presented here, form recognition could proceed from the bottom-up by first learning a sparse representation, and then by analyzing it to see if one or more forms fit sufficiently well. For cases such as the color circle where the sparse graph exactly instantiates a form (Fig. \ref{fig_main_set}B), form could be identified by simply checking the grammatical definition of that form (see sketch of such a mechanism in Fig. \ref{fig_form_laws}). For example, a ring could be identified as any connected graph with two edges per node. In fact, there is behavioral support for a related mechanism: people seem to be capable of identifying the abstract form of sparse relational graphs, when the edges are presented sequentially as social interactions \citep{Kemp2007,Kemp2008b}. Recognizing forms with exceptions (a tree with cross-branch relations) or forms with more complex structures (``a ring of trees'' or ``a tree in a ring'', Fig. \ref{fig_synthdata}B) may require probabilistic inference over a more sophisticated description language, such as probabilistic predicate logic \citep{Richardson2006} or probabilistic programs \citep{Goodman2008}. In any case, if sparsity is as powerful as our results suggest, this additional mechanism may provide a relatively modest step in facilitating the rare yet pivotal conceptual leaps in science and development. A criterion for judging any model of learning is how much richness is gained from initial knowledge to acquired knowledge. By this measure, sparsity could play the primary theoretical role in a larger explanation of qualitative structural change.

\begin{figure}[h]
\centering
$\begin{array}{ll}
\hline \text{Logical law} & \text{Interpretation} \\ \hline
1.\ \forall x \forall y\ \neg R(x,y) & \text{No edges between cluster nodes.}\\
2.\ \exists x \existsexactly y\ R(x,y) & \text{At least one cluster node has just one edge.} \\
3.\ \forall x \forall y\ R(x,y) & \text{Exactly 2 edges per cluster node.} \\
\ \ \ \ \ \ \to \existsexactly z [z \ne y \wedge R(x,z)]  \\
4.\ \forall x \forall y\ R(x,y) & \text{Exactly 1 or 2 edges per cluster node.} \\ 
\ \ \ \ \ \ \to ( \existsexactly z [z \ne y \wedge R(x,z)] \\
\ \ \ \ \ \ \vee\ \neg \exists z [z \ne y \wedge R(x,z)]) \\
5.\ \forall x \forall y\ T(x,y) & \text{At least one path between cluster nodes.} \\
6.\ \forall x \forall y \forall z \ [R(x,y) \wedge R(x,z) & \text{There are no cycles.} \\
\ \ \ \ \ \  \wedge\ y \ne z] \to \neg T_{\backslash x}(y,z) \\ \hline
\end{array}$
\caption{Sketch of an additional mechanism for identifying structural form by analyzing the output of the structural sparsity model \citep[inspired by ][]{Kemp2008b}. Forms are defined as conjunctions of laws that operate on the graph output: ``Clusters'' : Law 1; ``Chain'' : 2, 4 and 5; ``Ring'' : 3 and 5; ``Tree'' : 5 and 6. By applying these rules, the process can identify structural sparsity's output on the color circle as a ring (Fig. \ref{fig_main_set}B) and on mammals as a tree (Fig. \ref{mammals_graphs}A). The relation $R(\cdot,\cdot)$ indicates that two cluster nodes share an edge. Since edges are undirected, there is an implicit law of symmetry. The quantifiers are ``for all'' ($\forall$), ``there exists'' ($\exists$), and the non-standard ``there exists exactly one'' ($\existsexactly$). The predicate $T_{\backslash x}(\cdot,\cdot)$ is the transitive closure of $R(\cdot,\cdot)$ on the set of all objects excluding $x$.}
\label{fig_form_laws}
\end{figure}

\section{Acknowledgments}
We thank Intel Labs for providing the data set of 200 objects and Venkat Chandrasekaran for helpful discussions. This work was supported by a NSF Graduate Research Fellowship to B.M.L; the Center for Brains, Minds, and Machines funded by NSF Science and Technology Center award CCF-1231216; and the Moore-Sloan Data Science Environment at NYU. 

\section{Appendix} \label{appendix}

\subsection{Generating data from structure} \label{appendix_gen_model}
A data set is a matrix $D = \{f\hi{1},...,f\hi{m}\}$ ($D \in \mathbb{R}^{n_x \times m}$), where the rows correspond to objects and the columns are features $f\hi{i}$. A data set is generated by a structure, parameterized by a symmetric matrix $S \in \mathbb{R}^{n_t \times n_t}$. Each row/column in $S$ corresponds to a node in the graph. The set of object nodes is called $X$ and cluster nodes is called $Z$, where $n_x$ and $n_z$ are their cardinalities, combining for a total of $n_t = n_x + n_z$ nodes in the graph.

A key property of the structural sparsity model is that $S$ is sparse, meaning that most of its entries are equal to zero ($s_{ij} = 0$). For the remainder, their values are positive ($s_{ij} = s_{ji} > 0$). The sparsity pattern defines the adjacency matrix of the undirected graph, where a non-zero value for $s_{ij}$ means that nodes $i$ and $j$ share an edge. Given that each object node $x \in X$ connects to only one cluster node, then $s_{xz} > 0$ for exactly one $z \in Z$. There are no self-edges, so the diagonal of $S$ is also zero.

Following \citet{Zhu2003} and the setup for the structural forms model \citep{Kemp2008}, we introduce the graph Laplacian $\Delta$. Its off-diagonal elements are given by $-S$ and its diagonal elements are
\begin{equation*}
\Delta_{ii} = \sum_j s_{ij}.
\end{equation*}
A generative model for features can then be written as
\begin{eqnarray*}
P(f\hi{k} | S) & \propto & \exp\Big(-\frac{1}{4} \sum_{i,j} s_{ij} (f\hi{k}_i - f\hi{k}_j)^{2}\Big) \\
& = & \exp\Big(-\frac{1}{2} f^{(k)\top} \Delta f\hi{k}\Big).
\end{eqnarray*}
This equation highlights why the model favors features that are smooth across the graph. Features are probable when connected objects $i$ and $j$ ($s_{ij} > 0$) have a similar value. The stronger the connection $s_{ij}$ (meaning the larger its value), the more important it is for the feature values to match. As pointed out in \citet{Zhu2003}, this distribution is not proper, since adding a constant value to the feature vector does not change its probability. Therefore following \citet{Zhu2003}, we define the matrix $J = \Delta + \frac{1}{\sigma^2} I$ and use
\begin{equation} \label{eq_prec}
P(f\hi{k} | S, \sigma^2) \propto \exp\Big(-\frac{1}{2} f^{(k)\top} J f\hi{k}\Big),
\end{equation}
which results in a proper density. This distribution is an $n_t$ dimensional Gaussian with zero mean
\begin{eqnarray*}
P(f\hi{k} | S, \sigma^2) & = & N\Big(f\hi{k} | 0, J^{-1}\Big) \\
& = & \frac{1}{(2\pi)^{n_{t}/2} |J|^{-1/2}} \exp\Big(-\frac{1}{2} f^{(k)\top} J f\hi{k}\Big).
\end{eqnarray*}
This generative model for features can also be derived from a maximum entropy formulation, as shown in \citet{Lawrence2011,Lawrence2012}. This distribution over features is nearly the same as in the structural forms model, except the forms model adds the diagonal term $\frac{1}{\sigma^2}$ only to the observed nodes.

This distribution is also known as a Gaussian Markov Random Field \citep{Koller2009}. The undirected graph $S$ (and equivalently the precision matrix $J$) instantiates a set of conditional independence relationships between the variables in the graph
\begin{equation*}
s_{ij} = 0 \text{\ if and only if\ } \Big(f\hi{k}_i \perp f\hi{k}_j | f\hi{k}_{\backslash\{i,j\}}\Big).
\end{equation*}
If nodes $i$ and $j$ do not share an edge, their feature values are conditionally independent when the rest of the feature vector is observed (or equivalently, their partial correlation is 0 when controlling for all other nodes). Intuitively, a shared edge is a path of direct dependence.

Given that each object connects to exactly one cluster node (and no other nodes), there are no paths of direct dependence between objects. Instead, covariation between objects is represented through the connections between their cluster nodes. Moreover, two objects assigned to the same cluster node are conditionally independent, if the feature values of their shared cluster node could be observed.

\subsection{Prior distribution on structure}
The prior on structures $S$ has a very simple form
\begin{equation*}
P(S, \sigma^2) \propto \exp\Big(-\beta (\#S) \Big),
\label{sprior}
\end{equation*}
where $\#S$ is the number of edges in a structure. Equivalently, $\#S = \frac{1}{2} ||S||_0$, where $||S||_0$ known as the $\ell_0$-norm, counts the number of non-zero entires in a matrix. The prior has support on the set of all possible graphs in which the following conditions are met: each object has only one connection (which is to a cluster node), there are no empty cluster nodes, each matrix entry is non-negative ($s_{ij} \ge 0$), and $\sigma^2 > 0$.

This prior is improper because the values in $S$ have no upper bound. Given that this paper aims to just find a single good structure, this issue can be ignored and the prior becomes a penalty $-\beta(\#S)$ in the model's log-score (Eq. \ref{score}). If it was desirable (and tractable) to compute a posterior distribution $P(S|D)$, then this improper prior could be extended to form a proper prior. Following the structural forms model, $S$ could be decomposed into its sparsity pattern $S_{pat}$ (which values are non-zeros) and the edge values $S_{val}$. The forms model places a prior on each entry in $S_{val}$, and uses Laplace's method \citep{MacKay2003} to approximate the integral over the parameter values 
{\small\begin{equation*}
P(D|S_{pat}) = \int P(D|S_{pat},S_{val},\sigma^2) P(S_{val}|S_{pat}) P(\sigma^2) dS_{val} d\sigma^2.
\label{bayesian_score} 
\end{equation*}}
This strategy is entirely consistent with the structural sparsity model. In fact, approximating the integral would create an additional force for sparsity, known as the Bayesian Occam's razor \citep{MacKay2003}. But our current approach was chosen for simplicity, since we found it unnecessary to have two simultaneous forces driving for sparsity. There are some more formal reasons to support this point. As the number of features grows, the  Laplace approximation for the integral above asymptotes to a model score known as the Bayesian Information Criterion (BIC) \citep{Schwarz1978,Koller2009}. The BIC score maximizes the likelihood while penalizing the number of parameters, just like the penalization arising from our prior on $S$, although it differs in a scaling factor. Thus the more complex model, which combines these two sources of sparsity, asymptotes to the simpler model with just a larger value of the sparsity parameter $\beta$, soaking up both forces that promote sparsity.

\subsection{Model implementation} \label{appendix_search}
Computing a posterior distribution over structures $P(S|D)$ is very difficult, and thus we aim to find a single good structure $S$ that maximizes the posterior score (Eq. \ref{score}). Our approach to this optimization problem consists of two main routines. The outer-routine searches for the best clustering pattern (partition) of objects into cluster nodes, called \emph{cluster search}. The inner-routine searches for the best graph $S$ without changing the clustering pattern, called \emph{connection search}.

\subsubsection{Cluster search}
The strategy for cluster search is related to standard methods for structure learning in graphical models \citep{Koller2009}, which evaluate small changes to a graph and often greedily select the best change. Given a hypothesis structure and its cluster pattern, cluster search considers a set of local proposals to split or merge cluster nodes. The best scoring move, whether it is a split or a merge, is chosen greedily at each step. In order to assess the quality of each move, the edges must be re-optimized with the connection search routine which is computationally expensive. But multiple possible moves can be evaluated in parallel. To help mitigate the problem of local optima, search does not terminate immediately when the best local move decreases the score. Instead, the algorithm terminates after the score decreases several times (we used 5). Search also keeps a tabu list of previously visited cluster partitions that it does not revisit. The best evaluated structure $S$ is returned as the solution.

\textbf{Initialization}.
Choosing a good initialization can save a lot of computation, and in some cases, lead to better results. Given that cluster search operates over cluster partitions, we use standard clustering methods to propose a set of reasonable candidates for initialization. We use k-means clustering to choose a set of $k$ clusters. Multiple values of $k$ are tried, where the values of $k$ are evenly spaced on a logarithmic scale from 1 to the number of objects $n_x$. Each candidate is evaluated and scored with connection search. After determining the best value for $k$ in this initial coarse sampling, the algorithm attempts to narrow in on a better value. Picking the closest, previously-tried values of $k$ above and below the current $k$, $k$ is further optimized in this range by Fibonacci search. Assuming the score is a unimodal function of $k$ within these bounds, Fibonacci search will find the optimum in this range.

\textbf{Splitting clusters}.
To split a cluster node, the objects currently assigned to that node must be divided between the two new clusters. Following the general splitting strategy used in the structural forms algorithm \citep{Kemp2008}, the sparsity algorithm chooses two seed objects at random, assigns one to each cluster, and stochastically distributes the remaining objects by picking whichever seed object is closer in feature space. There is also a low probability of choosing the opposite seed object. Rather than evaluating just one split per node, the algorithm tries several (3) randomly chosen seed objects. Each split must then be optimized with connection search and scored by the objective function (Equation \ref{score}). For each step during search, the algorithm is limited in the total number of splits it will consider across all of the cluster nodes (30 for most data sets, but 8 for large data sets like the Senate and 200 Objects).

\textbf{Merging clusters}.
To merge two cluster nodes, they must combine their attached objects to form a single cluster node. Rather than trying every combination of merges, each cluster node stochastically, but intelligently, picks another to combine with. To help select the merges, the algorithm first calculates the expected value of the latent features for all cluster nodes, and the probability of merging two nodes decreases with the distances in this feature space. The algorithm also limits the number of merges it evaluates (30 for most data sets, but 8 for the Senate and 200 objects).

\textbf{Swapping assignments}.
In addition to splits and merges, the algorithm also tries to swap the cluster assignments of objects \citep{Kemp2008}. These moves do not compete with splitting and merging during each greedy step of the algorithm. Instead, at regular intervals throughout search (we used every 3 moves), each object tries to change its cluster assignment while leaving all others fixed. It tries all possibilities, but it does not re-learn the sparsity pattern for each possible assignment, which is an expensive computation. Instead, it just re-optimizes the existing edge strengths of each candidate proposal $S$. If a new parent leads to a better score, then $S$ is re-optimized with the full connection search.

\subsubsection{Connection search}
Connection search is the sub-routine that searches for the best sparse connectivity pattern, given a assignment of objects to cluster nodes. Defining a function $c(S)$ that extracts the cluster assignment, connection search must solve
\begin{equation}
\amax{\sigma^2,\{S: c(S) = w\}}  \sum_{i=1}^m \log P(f\hi{i}_X | S,\sigma^2) - \beta(\#S), \label{sem_obj}
\end{equation}
where the current cluster assignment is denoted as $w$. Given that features are only observed for object nodes, and even those features can be missing, we use the Structural Expectation-Maximization algorithm (Structural EM or SEM) for learning in the presence of missing data \citep{Friedman1997}.

\textbf{Structural Expectation-Maximization}.
SEM reduces the missing data problem to a sequence of simpler structure learning problems with complete data. Rather than maximizing Eq. \ref{sem_obj} directly, the structure at iteration $r+1$ of SEM maximizes
\begin{equation} \label{sem_obj_expectation}
S^{[r+1]},\sigma^{2[r+1]} \leftarrow \amax{\sigma^2,\{S: c(S) = w\}} \sum_{i=1}^m E_{Q_i(f_Z\hi{i}, S^{[r]},\sigma^{2[r]})}\Big[ \log P(f\hi{i}_X,f\hi{i}_Z | S,\sigma^2)\Big] -\beta(\#S),
\end{equation}
% \begin{multline*}
% S^{[r+1]},\sigma^{2[r+1]} \leftarrow \\
% \amax{\sigma^2,\{S: c(S) = w\}}
% -\beta(\#S) + \sum_{i=1}^m E_{Q_i(\cdot,[r])}[ \log P(f\hi{i}_X,f\hi{i}_Z | S,\sigma^2)] ,
% \end{multline*}
where the expectation is over the conditional distribution of the missing or latent features ($f_Z\hi{i}$) given the  observed features ($f_X\hi{i}$) and the current structure at iteration $r$,
$$ Q_i(f_Z\hi{i}, S^{[r]},\sigma^{2[r]}) = P(f_Z\hi{i} | f_X\hi{i}, S^{[r]},\sigma^{2[r]}).$$
Each iteration of the Structural EM algorithm is guaranteed to improve the original objective, or the marginal probability of the observed features Eq. \ref{sem_obj}. For the structural sparsity model, each iteration can be decomposed into an E-step (Step 3 below) which computes expected sufficient statistics. This is followed by a Structural M-step that re-optimizes the structure to fit the new sufficient statistics (Step 4). The algorithm is shown below.
\begin{algorithmic}[1] \label{SEM}
\STATE Initialize $S^{[0]}$ and $\sigma^{2[0]}$ \\
\FOR{ $r=0,1,2,...$ until convergence}
\STATE $H\  \leftarrow \frac{1}{m} \sum_{i=1}^m E_{Q_i(f_Z\hi{i},S^{[r]},\sigma^{2[r]})} [f\hi{i} f^{(i)T} ]$ 
\STATE $\{S^{[r+1]},\sigma^{2[r+1]}\}\ \leftarrow \amax{\sigma^2,\{S: c(S) = w\}}  \text{log\ } |J| - \text{tr}(H J) - \frac{\beta}{m}||S||_0$
\ENDFOR
\end{algorithmic}
The precision matrix $J$ is implicitly a function of $S$ and $\sigma^2$. Step 4 is intractable, as is the case with most structure learning problems, and we describe how we approximately solve it in the next section. Also, we run traditional EM to convergence at the end of each iteration $r$ of Structural EM \citep[see Alternating SEM-EM][]{Friedman1997}. Traditional EM is the same algorithm as SEM but with $\beta=0$ and a fixed sparsity pattern for $S$ across iterations.

\textbf{Optimization with $\ell_1$ heuristic}.
Exactly solving Step 4 is intractable, since all possible sparsity patterns need to be considered and the number of different sparsity patterns for $k$ clusters is $2^{(k^2-k)/2}$. Instead we use a convex relaxation of the optimization which replaces the $\ell_0$ norm with the $\ell_1$ norm. This heuristic leads to a sparse solution \citep{Banerjee2008}. The $\ell_1$ relaxation penalizes a sum of the parameters instead of their cardinality, such that $||S||_0 \approx  \lambda \sum_{ij} s_{ij}$ for some constant $\lambda$. This leads to a convex optimization problem which can be solved exactly \citep{Lake2010}:
\begin{equation*} \label{opt_prob}
\begin{aligned}
\amax{J,S,\sigma^2} 
& \text{log\ } |J| - \text{tr}(H J) - \frac{\beta \lambda}{m} \sum_{ij} s_{ij} \\
\text{subject to}\\
& J = \text{diag}\Big(\sum_j s_{ij}\Big) - S + \frac{1}{\sigma^2} I \\
& s_{ij} = 0, \; \{i,j\} \notin {\cal{E}} \\
& s_{ij} \ge 0, \; \{i,j\} \in {\cal{E}} \\
& \sigma^2 > 0.
\end{aligned}
\end{equation*}
The set of allowable edges is denoted as ${\cal{E}}$. Given that $J$ can be defined as an affine mapping from the variables $S$ and $\sigma^2$, we can make the first equality constraint implicit in the objective function and remove $J$ as a variable \citep{Boyd2004}. By setting the missing edges to be zero and removing them as variables, we have a convex optimization problem with just a positivity constraint on the parameters. Efficient methods for solving a related sparse Gaussian problem with $\ell_1$ have been developed \citep{Banerjee2008,Schmidt2009}, although this problem does not have a positivity constraint or the graph Laplacian interpretation. Algorithms for this related problem typically solve the dual optimization problem, although we found that solving the primal problem was more efficient for the variant defined here. To solve it, we used a projected quasi-Newton algorithm and code developed by \citet{Schmidt2009}. For a given solution, many values are zero but others are just small. The final sparsity pattern is chosen by thresholding to approximately maximize the complete-data objective in Step 4 of the Structural EM algorithm. The algorithm repeats the optimization and thresholding a total of three times, at $\lambda = \{.5, 1, 2\}$, and the best is picked according to the same objective. For the largest data sets, $\lambda = 1$.

\textbf{Derivation of Structural EM}.
More details of the Structural EM derivation are now described. We unpack the SEM objective function in Eq. \ref{sem_obj_expectation} and derive the algorithm in the section above. Here, the distribution $Q_i$ implicitly depends on the current hypothesis $S$, although we drop the dependence in the notation.

\bigskip
$\begin{array}{l}
\ \ \ \amax{\sigma^2,\{S: c(S) = w\}}  \log P(S,\sigma^2) + \sum_{i=1}^m E_{Q_i(f_Z\hi{i})} [\text{log\ } p(f_X\hi{i}, f_Z\hi{i} ; S,\sigma^2)] \\
%= \amax{\sigma^2,\{S: c(S) = w\}} \log P(S,\sigma^2)+  \sum_{i=1}^m E_{Q_i(f_Z\hi{i})} [-\frac{1}{2} \text{log\ } |J^{-1}|\ ...\\
%\ \ \ \ \ \ \ \ \ \ \ \ \ \ \ \ \ \   -\frac{1}{2} f^{(i)T} J f\hi{i} ] \\
= \amax{\sigma^2,\{S: c(S) = w\}} \log P(S,\sigma^2) -\frac{m}{2} \text{log\ } |J^{-1}|\ -\frac{1}{2} \sum_{i=1}^m E_{Q_i(f_Z\hi{i})} [f^{(i)T} J f\hi{i} ] \\
= \log P(S,\sigma^2)+  \frac{m}{2} \text{log\ } |J| -\frac{1}{2} \sum_{i=1}^m  (\sum_{j=1}^{n} \sum_{k=1}^{n} E_{Q_i(f_Z\hi{i})} [f\hi{i}_j f\hi{i}_k] J_{jk}) \\
= \log P(S,\sigma^2)+  \frac{m}{2} \text{log\ } |J|\ -\frac{1}{2} \sum_{i=1}^m \tr(E_{Q_i(f_Z\hi{i})}[f\hi{i} f^{(i)T}] J) \\
= \log P(S,\sigma^2)+  \frac{m}{2} \text{log\ } |J|  -\frac{m}{2} \tr(H J) \\
% = \amax{\sigma^2,\{S: c(S) = w\}} \frac{2}{m} \log P(S,\sigma^2)+  \text{log\ } |J| - \text{tr}(H J). \\
\propto  \text{log\ } |J| - \text{tr}(H J) - \frac{\beta}{m}||S||_0
\end{array}$

\medskip
\noindent 
Above, $J$ is the precision matrix (Eq. \ref{eq_prec}) and $H$ is the expectation of the empirical covariance,
$$ H  \leftarrow \frac{1}{m} \sum_{i=1}^m E_{Q_i(f_Z\hi{i})} [f\hi{i} f^{(i)\top} ].$$
The E-step (Step 3 in the SEM algorithm) uses this formula for computing $H$. Afterwards, we can approximately solve the optimization problem shown in the last line of the derivation, which forms the M-step (Step 4 in the SEM algorithm).

\bibliographystyle{apacite}
\bibliography{library_clean}
\end{document}